\pdfoutput=1

\documentclass[11pt]{article}

\usepackage{emnlp2021}

\usepackage{times}
\usepackage{latexsym}

\usepackage[T1]{fontenc}

\usepackage[utf8]{inputenc}

\usepackage{microtype}

\usepackage{amssymb}
\usepackage{amsmath}
\usepackage{booktabs}
\usepackage{threeparttable}
\usepackage{array}
\usepackage{float}
\newcolumntype{P}[1]{>{\centering\arraybackslash}m{#1}}
\usepackage{makecell}
\usepackage{multirow}
\usepackage{hhline}
\usepackage{subfig}
\usepackage[justification=justified]{caption}
\usepackage{graphicx}

\usepackage{amsthm}

\title{Improving Diversity of Neural Text Generation via \\Inverse Probability Weighting}

\author{Xinran Zhang\textsuperscript{1}, Maosong Sun\textsuperscript{12*}, Jiafeng Liu\textsuperscript{1}, Xiaobing Li\textsuperscript{1}\\
  \textsuperscript{1}Department of Music Artificial Intelligence and Music Information Technology \\Central Conservatory of Music, Beijing, China \\
  \textsuperscript{2}Department of Computer Science and Technology, Tsinghua University, Beijing, China\\
  Institute for Artificial Intelligence, Tsinghua University, Beijing, China\\
  State Key Lab on Intelligent Technology and Systems, Tsinghua University, Beijing, China\\
\texttt{zhangxr.wspn@gmail.com, sms@tsinghua.edu.cn}
  }
\date{}
\begin{document}
\maketitle
\begin{abstract}
The neural text generation suffers from the text degeneration issue such as repetition. Traditional stochastic sampling methods only focus on truncating the unreliable ``tail'' of the distribution, and do not address the ``head'' part, which we show might contain tedious or even repetitive candidates with high probability that lead to repetition loops. They also do not consider the issue that human text does not always favor high-probability words. Inspired by these, in this work we propose a heuristic sampling method. We propose to use interquartile range of the predicted distribution to determine the ``head'' part, then permutate and rescale the ``head'' with inverse probability. This aims at decreasing the probability for the tedious and possibly repetitive candidates with higher probability, and increasing the probability for the rational but more surprising candidates with lower probability. The proposed algorithm provides a reasonable permutation on the predicted distribution which enhances diversity without compromising rationality of the distribution. We use pre-trained language model to compare our algorithm with traditional methods. Results show that our algorithm can effectively increase the diversity of generated samples while achieving close resemblance to human text.
\end{abstract}
\section{Introduction}

Neural text generation is an important natural language processing (NLP) task, and have benefited a lot from Transformer ~\citep{NIPS2017_7181} architecture. However, it suffers from the well-known \emph{text degeneration} issue \citep{holtzman2019curious}, that is, the decoded texts exhibit a strong tendency to be repetitive with low diversity. To address this, many works have focused on stochastic sampling by truncating the ``\emph{tail}'' of the distribution, e.g., the top-\emph{k} sampling ~\citep{fan-etal-2018-hierarchical,holtzman-etal-2018-learning} or nucleus sampling (top-\emph{p} sampling, \citealp{holtzman2019curious}), which directly truncates the predicted distribution during sampling, excluding unreliable ``tail'' with low probability. Recent work by \citet{basu2021mirostat} adaptively truncates the ``tail'' to achieve controllable quality.


Regrettably, none of these methods have directly addressed the discrepancy that human text does \emph{not} always favor high-probability candidates ~\citep{holtzman2019curious}, i.e., the ``head'' of the distribution remains unprocessed. We show in our analysis that repetitive samples with low diversity are actually caused by the ``head'' part with high probability. Inspired by this, we propose the interquartile range inverse probability (IQR-IP) sampling algorithm. It brings a controllable permutation on the ``head'' part of the predicted distribution on the filtered vocabulary to enhance diversity without compromising the rationality of the distribution as well as the fluency of the generated text. Experiment results show that our algorithm can increase diversity while achieving close resemblance to human text compared with traditional methods.

\section{Observation on the ``Tail'' and ``Head''}

\subsection{Traditional Methods: Truncating the ``Tail'' to Balance between Quality and Diversity}

As is widely acknowledged for text generation, directly sampling on the predicted distribution will produce unsatisfactory samples due to the low-probability ``tail'' of the distribution (\citealp{holtzman2019curious}). This is self-explanatory since the ``tail'' contains unreasonable words that lead to less repetition as well as lower quality. Consequently, traditional methods always start by truncating the ``tail''. For example, the top-\emph{k} sampling ~\citep{fan-etal-2018-hierarchical,holtzman-etal-2018-learning} filters the top $k$ probable candidates from the vocabulary (denoted by $V$) as follows.
\begin{equation}\label{eq:topk}
V^{k}=\big\{x\mid rank\big(p(x)\big)\le k, x\in V\big\},
\end{equation}
where $p(x)$ denotes the predicted distribution of the language model, and $rank$ refers to the ranking order of $p(x)$. The auto-regressive dependency of $p(x)$ on the context of word $x$ on each sampling step is omitted for simplicity throughout this work.

According to ~\citet{holtzman2019curious}, top-\emph{k} sampling cannot address the discrepancy between \emph{peaked distribution} and \emph{flat distribution}. They propose nucleus sampling (top-\emph{p} sampling) which filters the vocabulary with top $p$ mass of cumulative probability as follows.
\begin{equation}\label{eq:topp}
V^{p}=\big\{x\mid cdf(x)\le p,x\in V\big\},
\end{equation}
where the cumulative density function $cdf(x)$ is calculated on the sorted distribution of $p(x)$. This produces better results than top-\emph{k} sampling, because it can dynamically drop more ``tails'' on peaked distribution, while top-\emph{k} sampling can't.


Clearly, these methods \emph{balance between quality and diversity by truncating the ``tail''}. Dropping more ``tails'' dynamically like nucleus sampling will improve quality but result in more repetition and lower diversity (see Table 1, ~\citealp{holtzman2019curious}), while keeping more ``tails'' like top-\emph{k} sampling will achieve less repetition but lower quality. Recent methods such as MIROSTAT by \citet{basu2021mirostat} adaptively truncate the ``tail'' with pre-defined quality target (perplexity) for better balancing effect.

\subsection{Repetition Loops Caused by the ``Head''}
\label{sec:head}
\begin{figure*}[hbt!]
\centering
\includegraphics[width=6in]{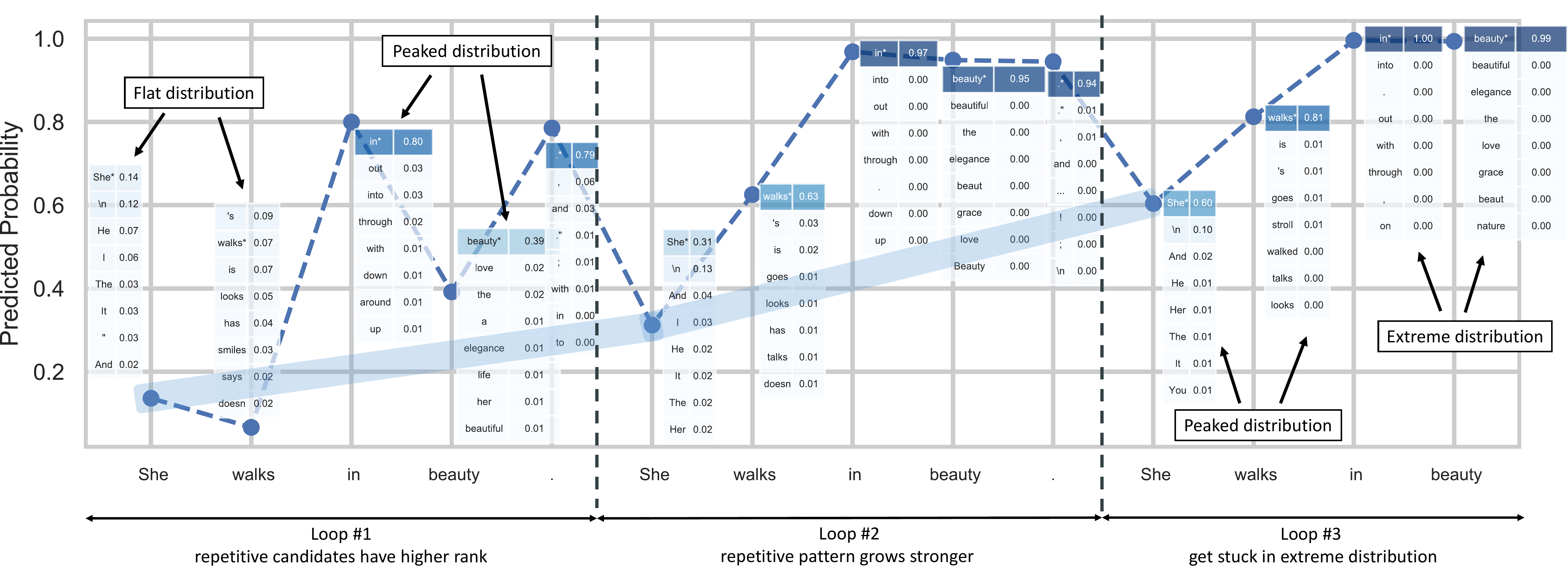}
\caption{Trajectory of predicted probability (``o'' marker) and predicted distribution (heatmap box besides each marker in ``word-probability'' format, with the sampled word marked by ``*'') for the first 3 repetition loops. This specific sample contains infinite repetitive loops of ``She walks in beauty.'' (with generated period). The trajectory of repetitive word ``She'' is highlighted in shadow which shows the increase of predicted probability and the gradually peaked predicted distribution.}\label{fig:head_repetitive_sentence}
\end{figure*}

\begin{figure*}[hbt!]
\centering
\includegraphics[width=6in]{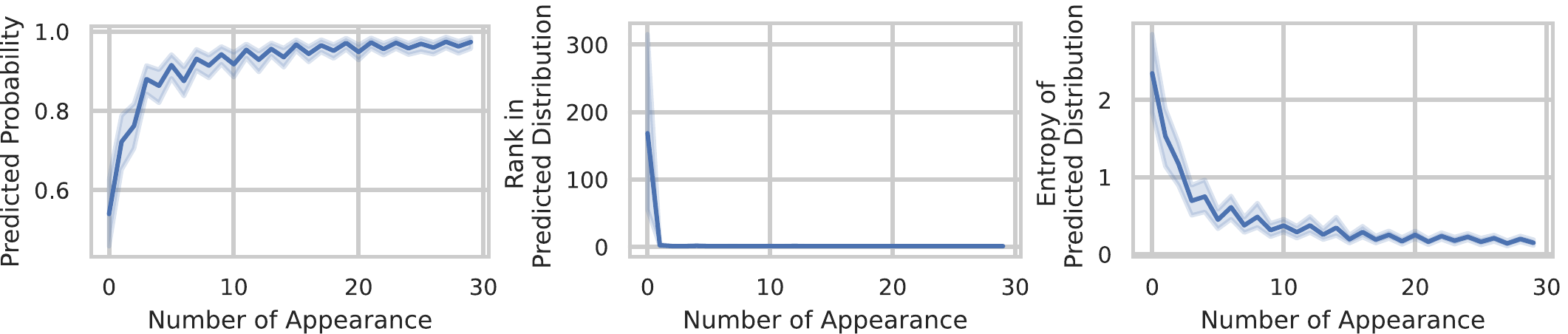}
\caption{Trajectories of repetitive candidates extracted from samples that contain repetition loops. Repetition loops are detected using $H_{rep}<2$ on 200-length token windows. Repetitive candidates that appears more than 30 times in the window are extracted and aligned to form their trajectories. It shows that a few appearances of repetitive candidates quickly lead the model to extreme distribution that causes repetition loops.}\label{fig:trajectory_rep}
\end{figure*}

However, traditional methods do not address the ``head'' part, which we show may lead to the annoying repetition loops.

To explore the behavior of repetition loops, we use GPT-2 Small ~\cite{radford2019language} with nucleus sampling ($p=0.95$) to generate 5,000 samples with the same input context and set maximum generation length to be 1,024. The sharing input context is ``She walks in beauty'' (from Lord Byron's most famous poetry).

To detect repetition as well as measuring the concentration tendency of vocabulary, we use a very straightforward metric by calculating the entropy of word distribution in a fixed-length window as follows.
\begin{equation}\label{eq:H_rep_1}
H_{rep} = -\sum_{w}{p(w)\times\log{p(w)}},
\end{equation}
\begin{equation}\label{eq:H_rep_2}
p(w) = f(w)/\sum_{w}{f(w)},
\end{equation}
where $f(w)$ denotes the frequency of word $w$. Samples with repetition loops will have concentrated distribution of $p(w)$ hence having lower $H_{rep}$, while samples with diverse usage of vocabulary will have flat distribution of $p(w)$ hence having higher $H_{rep}$. Empirically, we use $H_{rep}<2$ for all 200-length token windows to detect repetitive passages for observation.

We present a very representative sample that contains \emph{infinite loops} of ``She walks in beauty.'' (with generated period). The trajectory of first 3 generated loops is presented in Figure \ref{fig:head_repetitive_sentence}. We found several phenomena that cause this repetition.
\begin{itemize}
  \item Repetitive candidates always have \emph{high probability} and \emph{high rank} in the predicted distribution (see ``*'' labeled candidates in each heatmap box in Figure \ref{fig:head_repetitive_sentence}).
  \item Repetition tendency \emph{grows stronger} when \emph{more loops occur} (due to a few sampling steps that happen to pick repetitive token in non-extreme distribution, e.g, in Loop \#2), as the \emph{flat distribution} in Loop \#1 (e.g., ``She'' and ``walks'') gradually becomes \emph{peaked distribution} in Loop \#3, and peaked distribution in Loop \#1 (e.g., ``in'' and ``beauty'') becomes \emph{extreme distribution} in Loop \#3, which reciprocally contributes to stronger repetition pattern in the context.
  \item The predicted distribution got \emph{stuck} in \emph{extreme distribution} that assigns almost all probability mass for repetitive candidates (e.g., ``in'' and ``beauty'' in Loop \#3).
\end{itemize}

To further verify these phenomena, we extract and align the trajectories of each repetitive words to observe the overall trajectory for repetitive words (e.g., aligning all appearances of ``She'' sequentially on the $x$ axis). Figure \ref{fig:trajectory_rep} presents the trajectories of predicted probability, rank in predicted distribution and entropy of predicted distribution, where $x$ axis is the number of appearance of repetitive candidates. It shows that after a few appearances of repetitive candidates, the predicted distribution will quickly get stuck in extreme distribution where predicted probability approaches $1$, rank approaches $1$, and entropy approaches $0$, which will surely render repetition loops.

From these results, it is clear that the model tends to predict high probability for repetitive candidates that exist in the context. This is in accordance with analysis by ~\citet{kang-hashimoto-2020-improved}, which shows that words directly entailed in the context tend to have lower loss, i.e., higher predicted probability.

Clearly, these undesirable behaviors of the ``head'' with high probability will lead the model to generate samples that might contain repetition loops with low diversity. Regrettably, this issue is \emph{unable} to address by tradition stochastic sampling algorithms, since they still encourage to sample on high-probability candidates.

\subsection{Improving Diversity by Permutating the ``Head'' on Flat Distributions}

Recall the results by ~\citet{holtzman2019curious} which show that human text does \emph{not} always choose high-probability candidates, as the beam-search-based decoding method that generates samples with low perplexity actually \emph{deviates} from human text behavior (see Figure 2, ~\citealp{holtzman2019curious}). Our results in Section \ref{sec:head} also show that it will be \emph{harmful} to sample according to likelihood of candidates due to the behavior of the ``head''.

\begin{figure}[hbt!]
\centering
\includegraphics[width=3in]{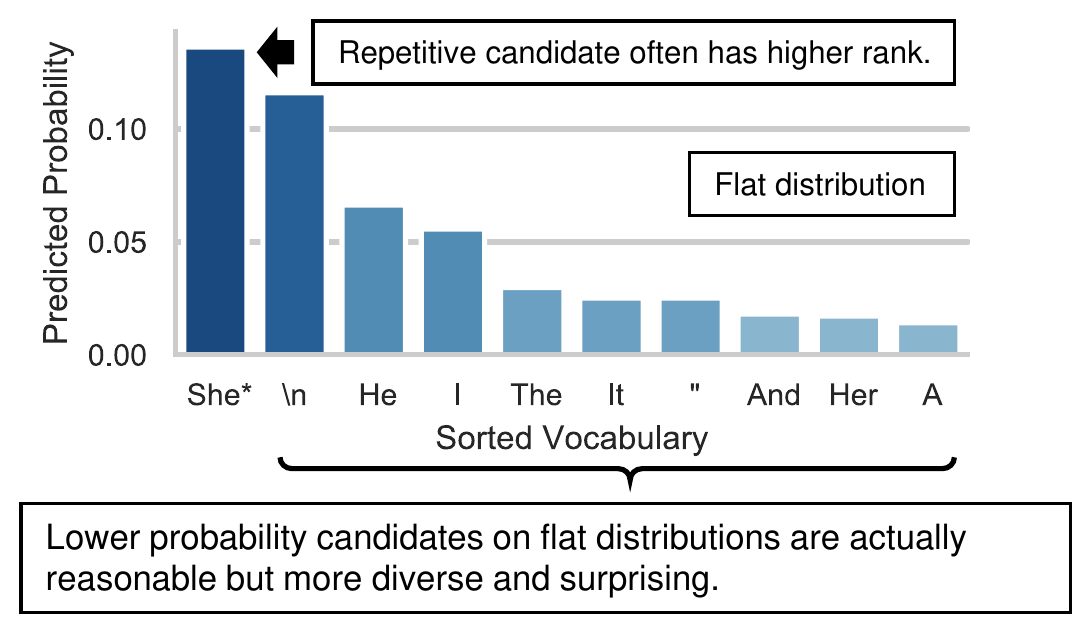}
\caption{Illustration of the ``Head'' on the flat distribution of the first sampling step of Loop \#1 from Figure \ref{fig:head_repetitive_sentence}. Besides ``She'' that has highest predicted probability, lower probability candidates (``\texttt{\textbackslash n}'', ``He'', ``I'', ``The'', ...) are also reasonable but more surprising with higher diversity. If using inverse probability weighting to emphasize on these candidates, the fluency of samples will not be compromised, while repetition will be suppressed and diversity will be improved.}\label{fig:head_distribution}
\end{figure}

To fix this, we present a detailed observation of the ``head'' in Figure \ref{fig:head_distribution}. It shows that \emph{lower-probability candidates} on a \emph{flat distribution} are actually \emph{reasonable} but \emph{more surprising} with \emph{higher diversity}. Consequently, it is possible to increase diversity by emphasizing on \emph{less probable} candidates on flat distributions without compromising the rationality of the distribution as well as the fluency of the generated text.

Intuitively, this can be achieved similarly to the \emph{inverse probability weighting} technique that is commonly seen in causal inference (see Chapter 2, ~\citealp{CausalInfBook}). Inspired by this, as long as we can identify a small subset of candidates (i.e., the ``head'') of the distribution that contains all reasonable candidates (such as in Figure \ref{fig:head_distribution}), we may use inverse probability weighting to rescale the distribution for these candidates to suppress repetition and increase diversity without compromising fluency.

\section{Interquartile Range Inverse Probability Sampling Algorithm}
\label{sec:3}

\subsection{Use Interquartile Range to Identify the ``Head''}

Clearly, the major difficulty in identifying the ``head'' is the variation of the shape of the distribution, i.e., the discrepancy between flat distribution and peaked distribution. Intuitively, the \emph{interquartile range} (IQR) can adapt to such variation since it is based on quantile calculation and does not have strict requirements for the shape of the distribution.

As a result, we propose to adopt IQR to identify the ``head'' for permutation. First, we need to ensure that only the most reliable candidates are kept in order not to interfere with the identification of the ``head''. Following the common filtering method of stochastic sampling, we propose to jointly filter an initial subset $V^{K_0}$ of candidates with $p$ and $k$ as follows.
\begin{equation}\label{eq:k_and_p}
V^{K_0} = V^{k}\cap V^{p}.
\end{equation}
Let $p_{fil}(x)$ denote the regularized distribution on $V^{K_0}$. We propose to calculate IQR of $p_{fil}(x)$, that is, calculate 75\% percentile as $Q_3$, 25\% percentile as $Q_1$, $IQR=Q_3-Q_1$, and divide $V^{K_0}$ into subsets as follows.
\paragraph{IQR Subset Division of $V^{K_0}$:}
\begin{equation}\label{eq:iqr_div}
\small
\begin{aligned}
V^{VeryHigh}:&\,\,p_{fil}(x)\ge Q_3+\rho\times IQR\\
V^{High}\,\,\,\,:&\,\,Q_3+\rho\times IQR>p_{fil}(x)\ge Q_3\\
V^{Medium}\,:&\,\,Q_3>p_{fil}(x)\ge Q_1\\
V^{Low}\,\,\,\,\,:&\,\,Q_1>p_{fil}(x)\ge Q_1-\rho\times IQR\\
\end{aligned},
\end{equation}
where $\rho$ is the hyper parameter for the coefficient of IQR with typical value being $1.5$. Considering the outlier-identification nature of IQR, $V^{VeryHigh}$ can be regarded as the ``head'' part that we need to permutate, which we expect that the least probable candidate in $V^{VeryHigh}$ is still likely to be ``high enough'' to be reasonable choices.

Since IQR is based on quantile, $V^{VeryHigh}$ is empirically to be \emph{non-singleton} on flat distribution only, hence permutation on $V^{VeryHigh}$ will \emph{not} interfere with peaked distribution which may compromise rationality of the distribution. See Appendix \ref{sec:explanation} for more discussions.

\subsection{``Leakage'' of the ``Tail'' on Peaked Distribution Interferes with Identification of the ``Head''}
\label{sec:leakage}
\begin{figure}[hbt!]
\centering
\includegraphics[width=2.8in]{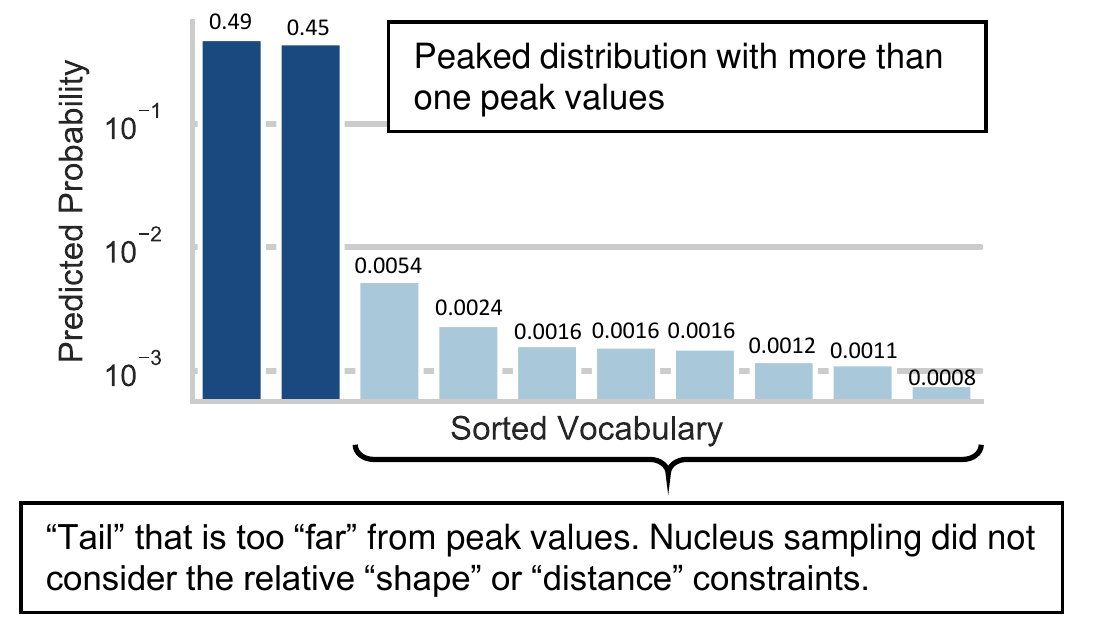}
\caption{The incurring of ``leakage'' of the ``tail'' on peaked distribution that has more than one peak value. On such distribution, small value of $p$ for nucleus sampling will miss the second peak, while large value of $p$ will easily let in low-probability candidates close to the peak (i.e., ``leakage''), which will interfere with the identification of the ``head''. This distribution is also selected from one of the generated samples using GPT-2 Small model.}\label{fig:tail_distribution}
\end{figure}
Before proceeding, we take a deeper look for ``tail'' part on peaked distribution. As is studied by ~\citet{holtzman2019curious}, nucleus sampling can adaptively truncate low-probability ``tails'' on peaked distribution, while top-\emph{k} sampling can't (see Figure 5, ~\citealp{holtzman2019curious}).

We consider a special case which is \emph{not} considered by ~\citet{holtzman2019curious} . Figure \ref{fig:tail_distribution} presents an actual example of peaked distribution with more than one peak value. In this case, small value of $p$ for nucleus sampling will miss the second peak, while large value of $p$ will easily let in low-probability candidates, i.e., resulting in ``leakage''. Although such leakage might affect very little on sampling (since the ``leakage'' part has low probability), but clearly it will affect the identification of ``head'' (since IQR calculation is based on quantile), hence cannot be ignored.

We argue that the incurring of leakage is because neither top-\emph{k} sampling nor nucleus sampling considers the relative ``shape'' or ``distance'' constraints during filtering. To fix this, we propose a new filtering metric to further exclude low-probability candidates  that is too ``far'' from the peaked ones. We define a threshold that is the fraction of the maximum probability on a predicted distribution, and exclude candidates with probability below that threshold, which we name as the ``top-1 controlled'' (\emph{top1ctrl}) filtering metric with parameter $n$ as follows.
\begin{equation}
V^{n}=\big\{x\mid p(x)\ge\max{p(x)}/n,x\in V\big\}.
\end{equation}
We propose to use this metric to prune $V^{K_0}$ (on the basis on joint vocabulary filtering in Equation \ref{eq:k_and_p}) in a dynamic way. Our method is described in the following equations, in which we denote the pruned set to be $V^{K_1}$.
\begin{equation}\label{eq:k1a}
\small
V^{K_1}=\left\{\begin{aligned}
&V^{VeryHigh}\cup V^{High},\quad\quad\text{if } V^{n}\subseteq \\ &\quad\quad\quad\quad\quad\quad \big( V^{VeryHigh}\cup V^{High}\big)\\
&V^{K_0}\cap V^{n},\quad\quad\quad\quad\quad\,\text{otherwise}\\
\end{aligned}\right..
\end{equation}
The first sub-equation ensures that $V^{n}$ does not truncate any candidates categorized as ``Very High'' or ``High'', since they are identified by IQR and likely to contain rational candidates. In this case we drop all candidates in $V^{Medium}$ and $V^{Low}$, because they are considered too ``far'' from maximum value in the distribution. And the second sub-equation describes other cases where $V^{n}$ works jointly with $V^{k}$ and $V^{p}$ in a straight-forward way. Practically $n$ is set to a fairly loose value of $100$ in our experiment in order to function correctly with top-\emph{k} filtering and nucleus filtering and not to over-prune $V^{K_0}$.

\subsection{Inverse Probability Permutation on the ``Head''}


With $V^{K_1}$ acquired, we propose to re-assign probability mass for each candidate in $V^{VeryHigh}$ (i.e., the ``head'') proportionally to its inverse probability, while keeping the sum of probability mass in $V^{VeryHigh}$ constant. In this way, distribution of the ``head'' is rescaled and has inverse monotonicity, while distribution on $V^{K_1}$ still maintains the probability distribution feature. For simplicity, now let $p_{fil}(x)$ denote the regularized distribution on $V^{K_1}$. The permutation on $V^{VeryHigh}$ is described as follows.
\begin{equation}\label{eq:iqr}
\small
\begin{aligned}
&p_{inv}(x)=\\
&\Bigg(\sum_{x\in V^{VeryHigh}}{p_{fil}(x)}\Bigg)
\times\frac{p_{fil}(x)^{-1}}{\sum_{x\in V^{VeryHigh}}{p_{fil}(x)}^{-1}},
\end{aligned}
\end{equation}
where $p_{inv}(x)$ denotes the permutated distribution, and $p_{inv}(x)$ outside $V^{VeryHigh}$ remains the same as $p_{fil}(x)$. Finally the stochastic sampling is performed according to $p_{inv}(x)$. We refer to the above algorithm as the \emph{interquartile range inverse probability} (IQR-IP) sampling algorithm. We summarize the main differences of our algorithm as follows.
\begin{itemize}
  \item We use dynamic vocabulary filtering with 3 parameters (\emph{p}, \emph{k}, and \emph{n}). This aims at guaranteeing the correct identification of the ``head'' of the distribution.
  \item Distribution of the ``head'' identified by IQR is permutated using Equation \ref{eq:iqr}. This aims at improving diversity by decreasing the probability of tedious and possibly repetitive candidates with high probability and increasing the probability of reasonable but more surprising candidates with low probability.
\end{itemize}

\subsection{Total Variance Analysis}

We provide total variance analysis to explain the behavior of our algorithm. Following proposition by ~\citet{kang-hashimoto-2020-improved}, we can evaluate the permutation by analyzing the upper bound of total variance between $p_{inv}(x)$ and reference distribution $p_{ref}(x)$ with the following corollary.
\paragraph{Corollary 1.} Upper bound of total variance between $p_{inv}$ and $p_{ref}$ satisfies
\begin{equation}\label{eq:proposition1-1}
\small
  |p_{inv}-p_{ref}|^2\le\frac{1}{2}KL(p_{ref}||p_{fil})+2m+m^2,
\end{equation}
where
\begin{equation}\label{eq:proposition1-2}
  m=\max_{x\in V^{VeryHigh}}{|p_{fil}-\frac{Z_{p}}{p_{fil}}|},
\end{equation}
\begin{equation}\label{eq:proposition1-3}
  Z_{p}=\frac{\sum_{x\in V^{VeryHigh}}{p_{fil}}}{\sum_{x\in V^{VeryHigh}}{p_{fil}}^{-1}}.
\end{equation}
See Appendix \ref{sec:appendix_proof} for proof.

Equation \ref{eq:proposition1-1} reveals an additional term controlled by $m$ besides the original bound $\frac{1}{2}KL(p_{ref}||p_{fil})$ (achieved by $p_{fil}$ without inverse probability permutation). Since $m$ contains an value of inverse probability, the new upper bound will change dramatically. This provides a \emph{controllable} diversity enhancement measure. See Appendix \ref{sec:ablation} for more analysis.
\def\MyWidth{0.085\textwidth}
\begin{table*}[hbt!]
\tiny
\centering
\tabcolsep 1mm
\renewcommand{\arraystretch}{1.4}
\begin{center}
\begin{tabular}{
P{0.08\textwidth}
P{0.13\textwidth}P{\MyWidth}P{\MyWidth}P{\MyWidth}P{\MyWidth}P{\MyWidth}P{\MyWidth}P{\MyWidth}P{\MyWidth}
}
\toprule
\multirow{2}{*}[-0.5em]{Model} & \multirow{2}{*}[-0.5em]{Method} & \multicolumn{5}{c}{Statistical Evaluation} & \multicolumn{3}{c}{Human Evaluation} \\
\cmidrule(l){3-7}\cmidrule(lr){8-10}& & PPL & {\tiny Self-BLEU 4} & {\tiny Self-BLEU 5}  & Zipf Coef. & $H_{rep}$ & Fluency $\uparrow$ & Diversity $\uparrow$ & Overall $\uparrow$\\
\midrule
\multirow{4}{*}[-2em]{\tiny GPT-2 Small}
& Human & 29.41 & 0.31 & 0.17 & 0.93 & 4.50 &  - & - & -\\
\cmidrule{2-10}
& \makecell{Nucleus, $p=0.9$} & \textbf{30.64} & \textbf{0.42} & \textbf{0.26} & 1.27 & 4.55  & 3.78 & 4.44 & 4.11\\
& \makecell{Top-\emph{k}, $k=200$} & 25.14 & 0.46 & 0.29 & 1.24 & \textbf{4.52}  & 3.74 & 4.56 & 4.15\\
& \makecell{IQR-IP (\emph{ours})\\$p=0.8, k=640$} & 32.88 & 0.43 & 0.27 & \textbf{1.05} & 4.61 & \textbf{3.87} & \textbf{4.64} & \textbf{4.25}\\
\midrule
\multirow{4}{*}[-2.5em]{\tiny GPT-2 XL}
& Human & 18.34 & 0.31 & 0.17 & 0.93 & 4.50 &  - & - & - \\
\cmidrule{2-10}
& \makecell{Nucleus, $p=0.9$} & 17.09 & \textbf{0.44} & 0.28 & 1.49 & 4.42  & 4.61 & 4.53 & 4.57\\
& \makecell{Top-\emph{k}, $k=200$}  & \textbf{17.86} & \textbf{0.44} & \textbf{0.27} & 1.41 & \textbf{4.45} & 4.56 & 4.64 & 4.60\\
& \makecell{IQR-IP (\emph{ours})\\$p=0.8, k=640$} & 16.77 & 0.47 & 0.29 & \textbf{1.17} & \textbf{4.45}  & \textbf{4.64} & \textbf{4.70} & \textbf{4.67}\\
\bottomrule
\end{tabular}
\end{center}
\caption{Statistical evaluation (\emph{closer metric to human text is better}) and human evaluation (\emph{higher score is better}) for selected decoding parameters. Note that our algorithm can achieve human level PPL with less repetition (with high $H_{rep}$). Also note the Zipf coefficient of our algorithm is \emph{much closer} to human metric and \emph{unable to achieve} by traditional methods. Human evaluation shows that our algorithm can achieve similar fluency but higher diversity.}\label{tab:main_result}
\end{table*}

\begin{figure*}[hbt!]
\centering
\captionsetup[subfigure]{justification=centering}
\subfloat[Perplexity for GPT-2 Small. Horizontal line (29.41, ~\citealp{radford2019language}) refers to human text.]{\includegraphics[width=1.2in]{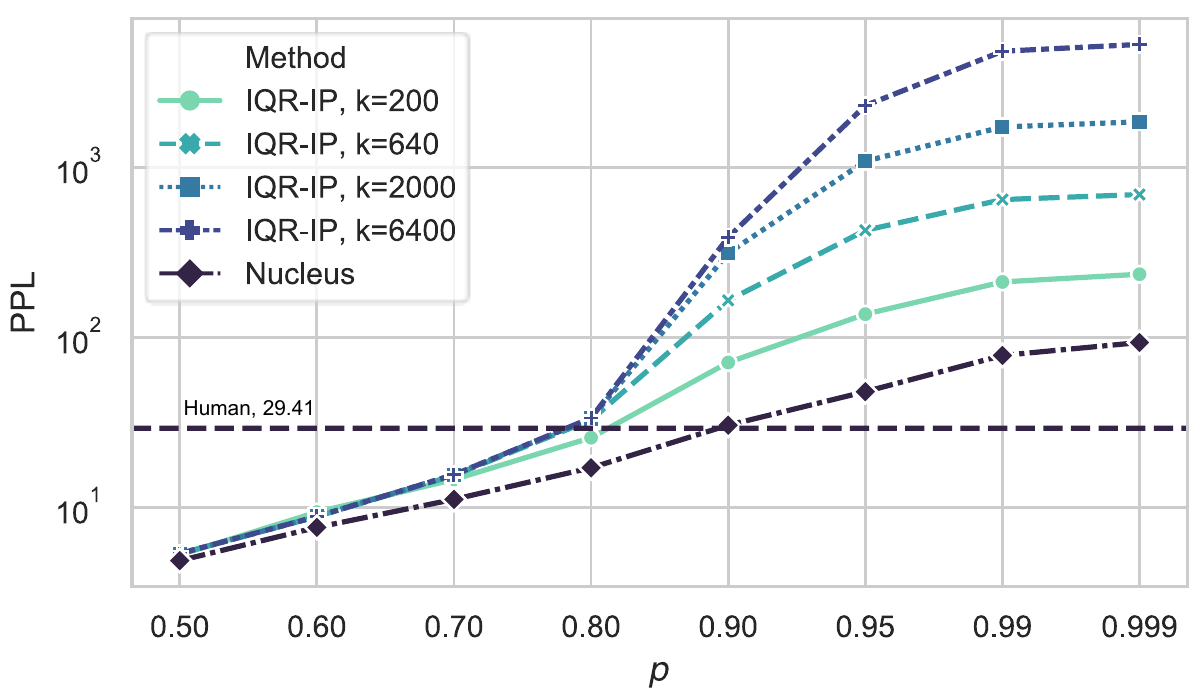}\label{fig:ppl_gpt2}}
\subfloat[Self-BLEU 4 for GPT-2 Small. Horizontal line (0.31, ~\citealp{{holtzman2019curious}}) refers to human text.]{\includegraphics[width=1.2in]{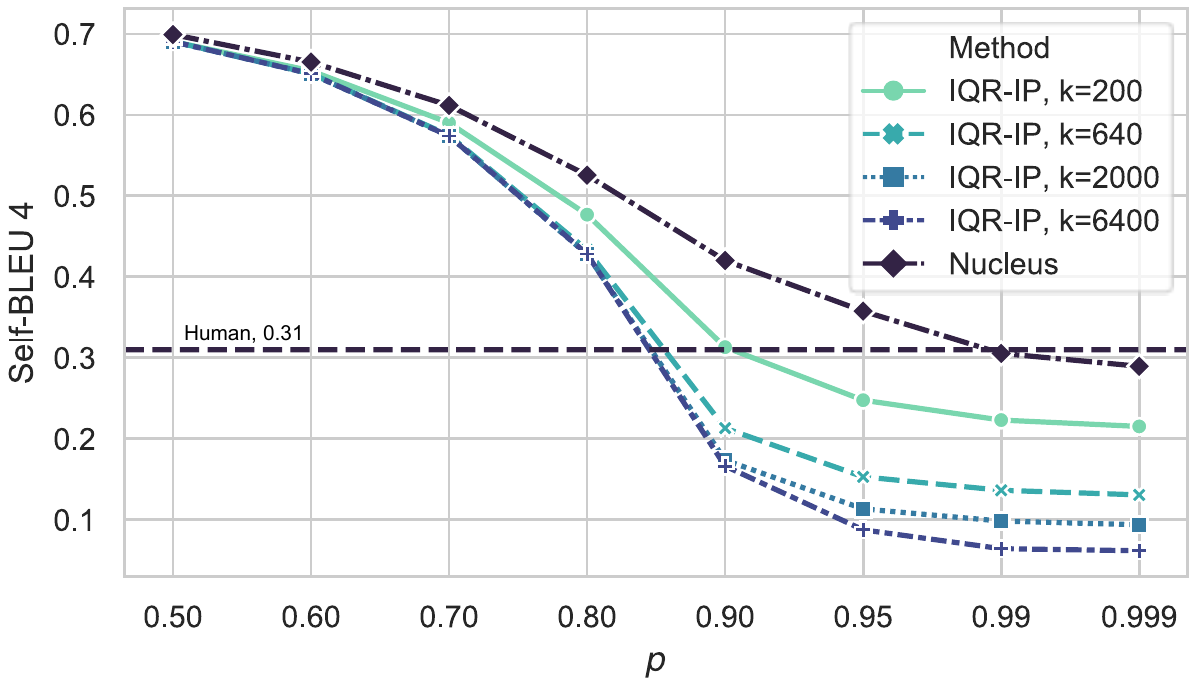}\label{fig:bleu4_gpt2}}
\subfloat[Self-BLEU 5 for GPT-2 Small. Horizontal line (0.17, ~\citealp{{holtzman2019curious}}) refers to human text.]{\includegraphics[width=1.2in]{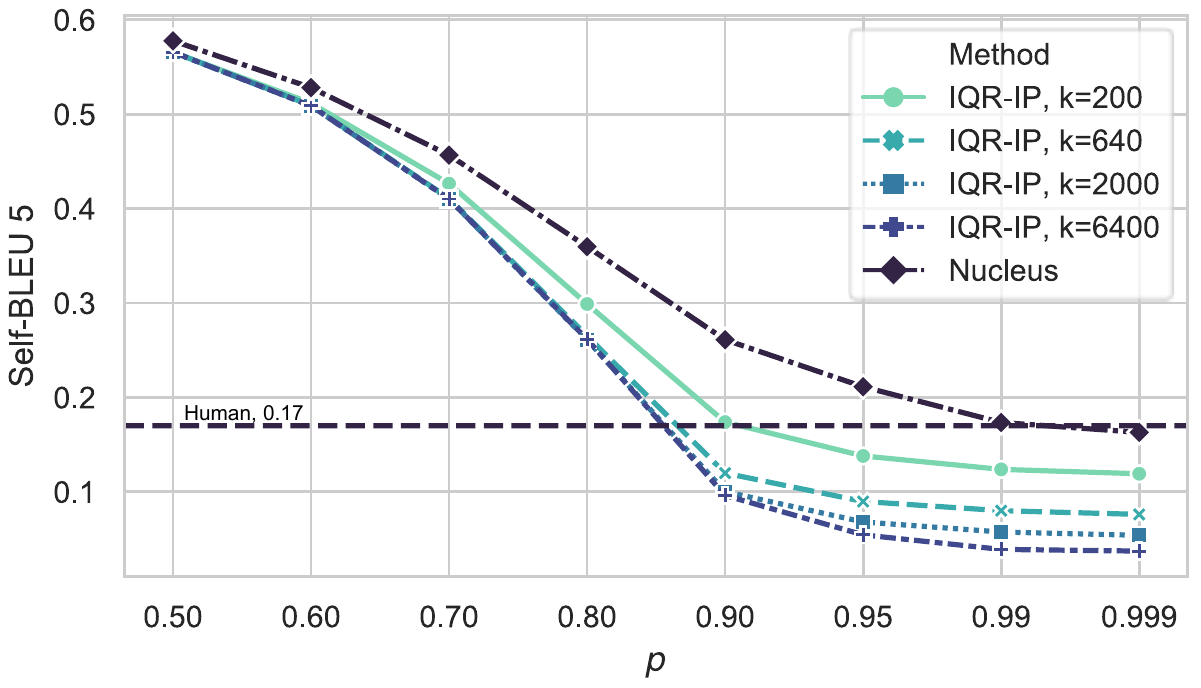}\label{fig:bleu5_gpt2}}
\subfloat[Zipf coefficient for GPT-2 Small. Horizontal line (0.93, ~\citealp{holtzman2019curious}) refers to human text.]{\includegraphics[width=1.2in]{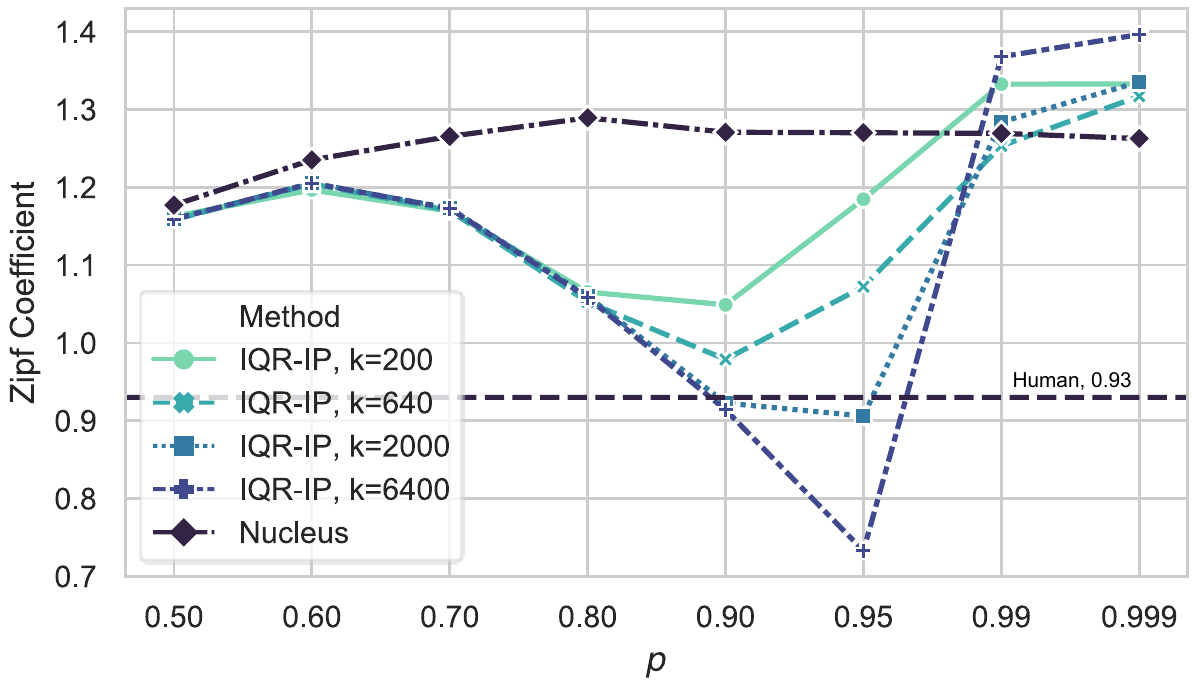}\label{fig:zipf_gpt2}}
\subfloat[$H_{rep}$ for GPT-2 Small. Horizontal line (4.50) refers to human text on test set of WikiText-2]{\includegraphics[width=1.2in]{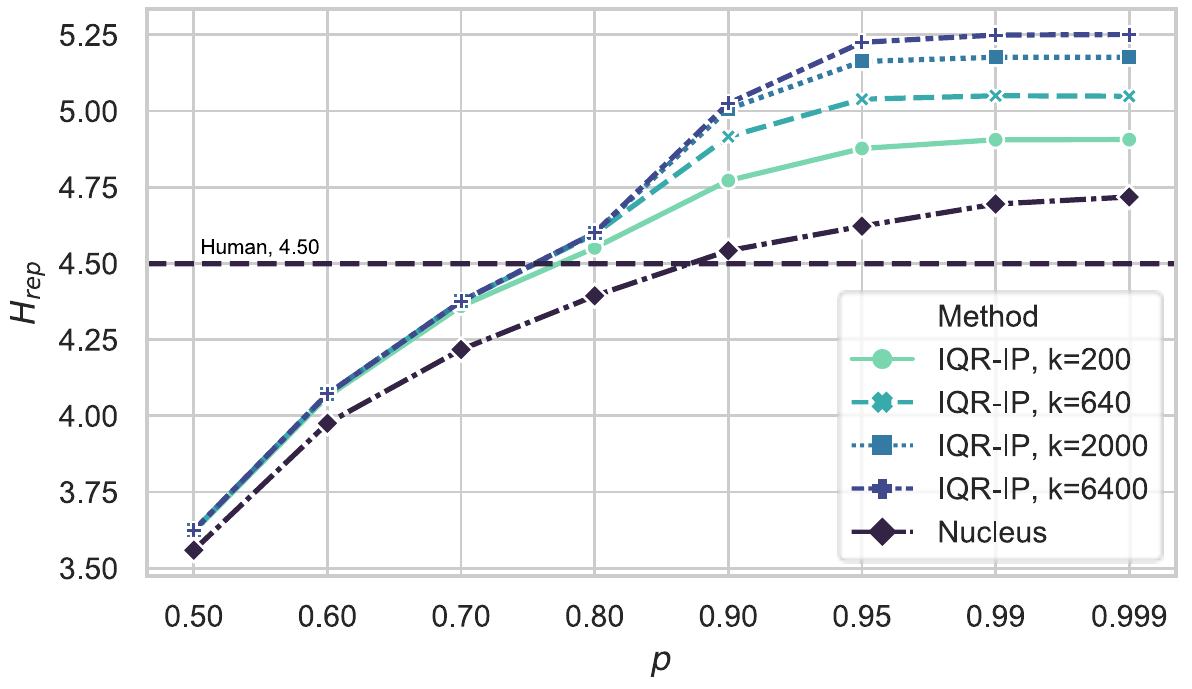}\label{fig:entropy_gpt2}}

\subfloat[Perplexity for GPT-2 XL. Horizontal line (18.34, ~\citealp{radford2019language}) refers to human text.]{\includegraphics[width=1.2in]{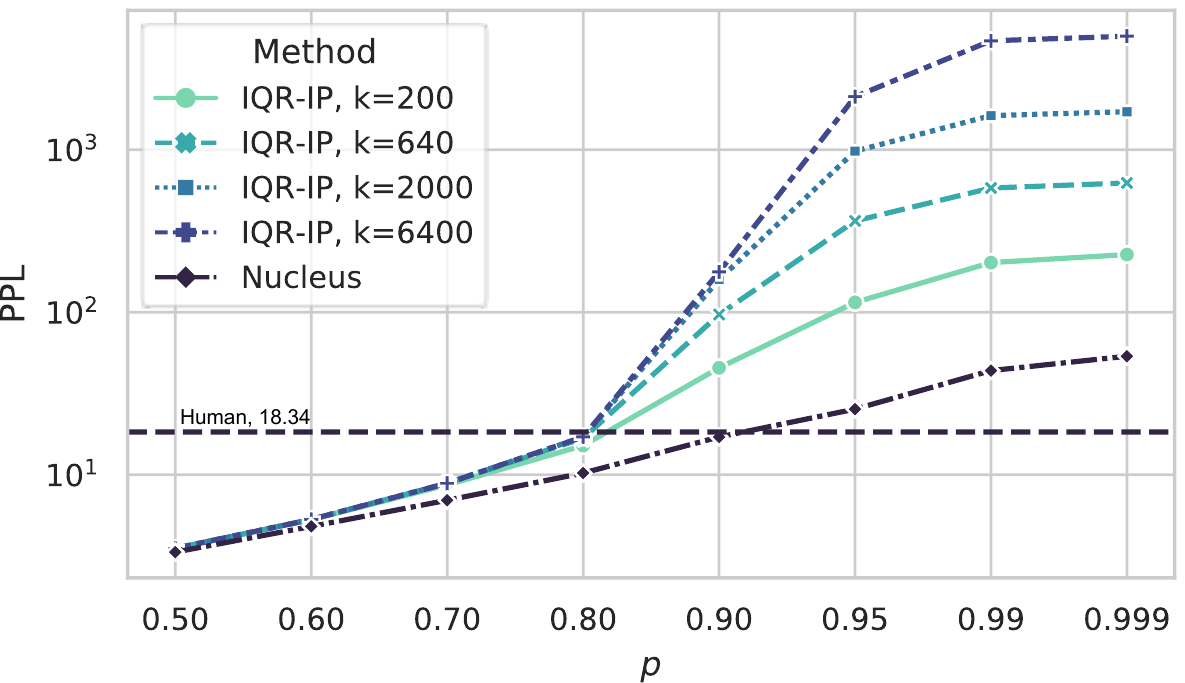}\label{fig:ppl_gpt2_xl}}
\subfloat[Self-BLEU 4 for GPT-2 XL. Horizontal line (0.31, ~\citealp{{holtzman2019curious}}) refers to human text.]{\includegraphics[width=1.2in]{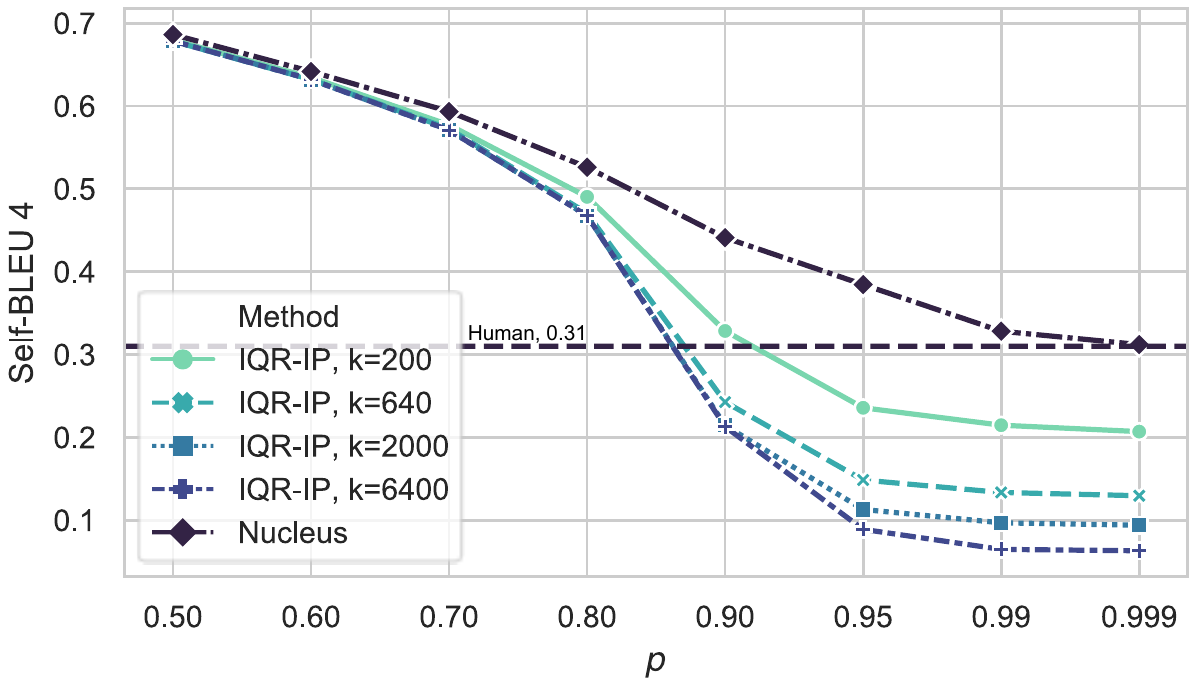}\label{fig:bleu4_gpt2_xl}}
\subfloat[Self-BLEU 5 for GPT-2 XL. Horizontal line (0.17, ~\citealp{{holtzman2019curious}}) refers to human text.]{\includegraphics[width=1.2in]{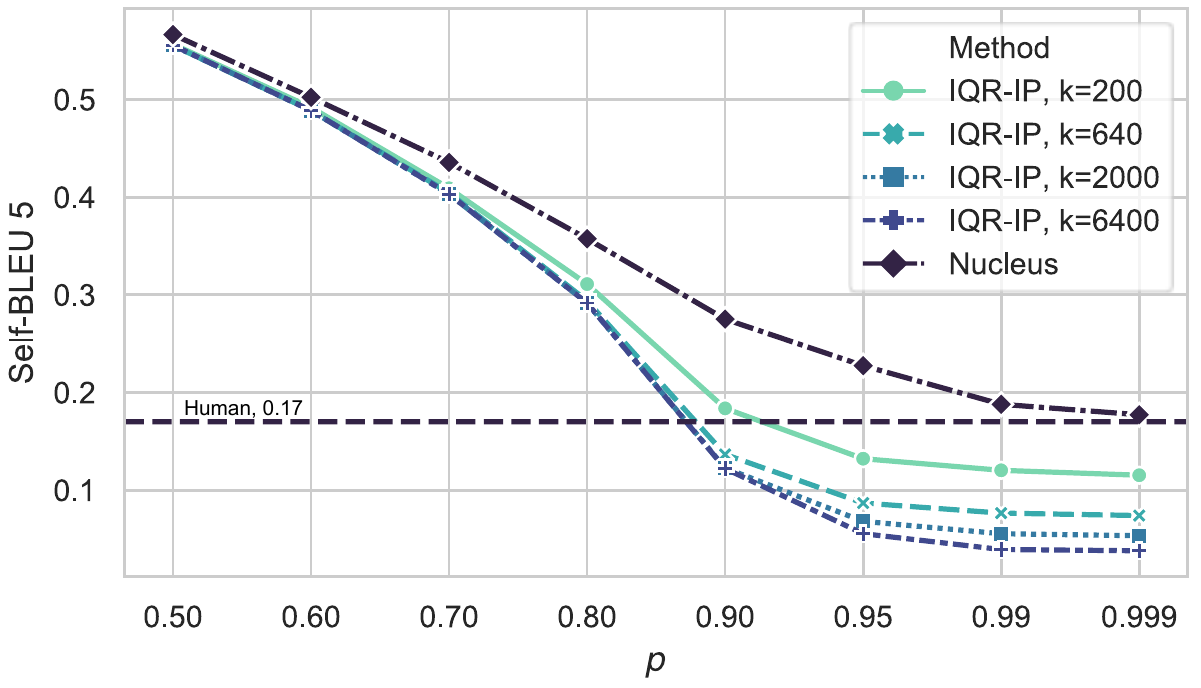}\label{fig:bleu5_gpt2_xl}}
\subfloat[Zipf coefficient for GPT-2 XL. Horizontal line (0.93, ~\citealp{holtzman2019curious}) refers to human text.]{\includegraphics[width=1.2in]{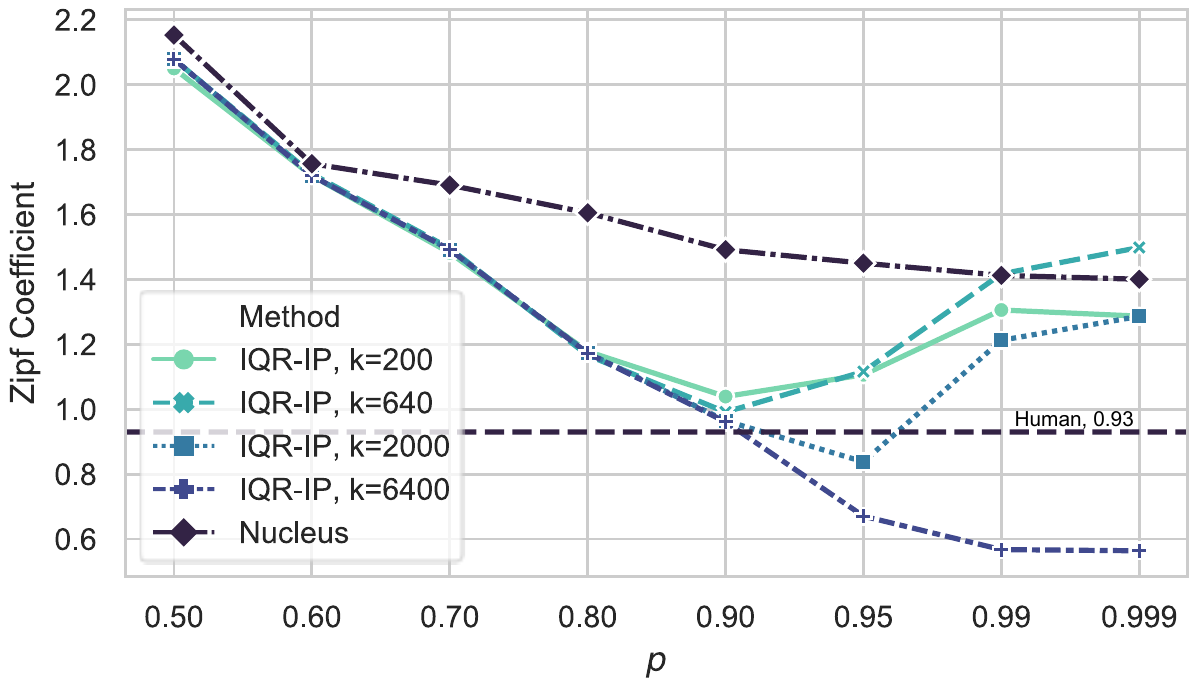}\label{fig:zipf_gpt2_xl}}
\subfloat[$H_{rep}$ for GPT-2 XL. Horizontal line (4.50) refers to human text on test set of WikiText-2]{\includegraphics[width=1.2in]{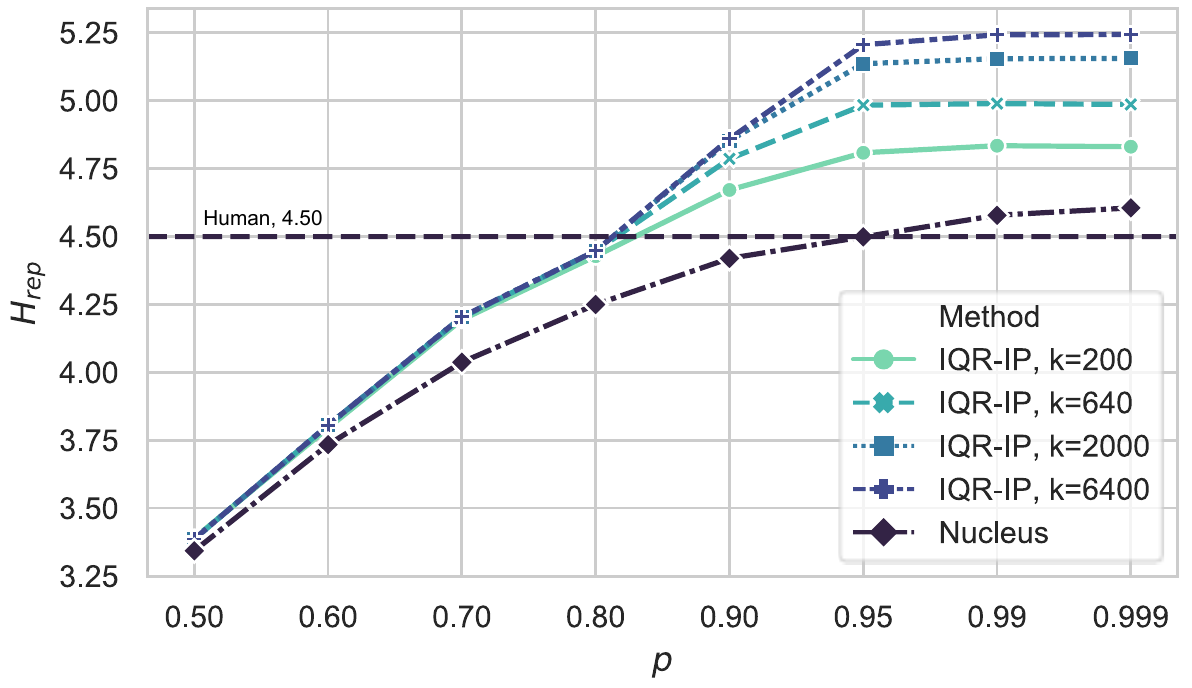}\label{fig:entropy_gpt2_xl}}
\caption{Statistical results and metric behavior comparison with nucleus sampling. They show that our algorithm achieves human level metrics with \emph{more strict} filtering parameters (i.e., with less ``tail''), which is contributed by the diversity gain from inverse probability permutation on the ``head''. Also note that the behavior of Zipf coefficient of our algorithm (with intersection to human metric) is significantly different from nucleus sampling (without intersection), because our algorithm encourages to sample on \emph{less probable} tokens of the ``head'' which renders more flat distribution of the vocabulary and achieves closer resemblance to human text.}\label{fig:all}
\end{figure*}
\section{Evaluation}
\label{sec:experiment}
\subsection{Experiment Setup}
The primary goal of the evaluation is to test whether our methods generate fluent samples with higher diversity. We consider the following principles when choosing baselines.
\begin{itemize}
  \item \textbf{Ablation of permutating the ``head''}. This means the baseline method should be \emph{without permutation}, i.e., choosing plain stochastic sampling that only truncates the ``tail'' for comparison.
  \item \textbf{Fair comparison on human-level PPL}. This means the baseline method as well as our method should already achieve close PPL to human text like Figure 6 by \citet{holtzman2019curious}, i.e., choosing hyper parameters \emph{near the intersection points with human PPL} for fair comparison.
  \item \textbf{Impact of model size}. This answers the question that does model size affect the conclusion from our experiments. We choose the \emph{smallest and largest} plain auto-regressive Transformer language models from GPT-2 family for interpolative conclusions.
\end{itemize}

We use pre-trained GPT-2 Small (117M parameters) and GPT-2 XL (1,542M parameters) released by ~\citet{Wolf2019HuggingFacesTS}. Following identical settings by ~\citet{holtzman2019curious}, we set maximum length of generation to be 200 and generate 5,000 samples for each sampling method with the same context in Section \ref{sec:head}. We set fixed value of $n=100$ for \emph{top1ctrl} filtering and $\rho=1.5$ for IQR.

\subsection{Statistical Evaluation}

We first follow the statistical evaluation procedure by ~\citet{holtzman2019curious}, which evaluates the following metrics (closer score to the human metric is better).
\begin{itemize}
  \item \textbf{Perplexity}. This metric is calculated on the \emph{generated texts} with the per-trained model to reflect its \emph{general quality and fluency}. Lower score indicates higher quality.
  \item \textbf{Self-BLEU (4 and 5)} \citep{holtzman2019curious,10.1145/3209978.3210080}. One sample is calculated against all other samples to reflect \emph{diversity among all samples}. Lower score indicates higher diversity.
  \item \textbf{Zipf coefficient} \citep{Zipf49,newman2005power}. This metric represents \emph{linguistic feature} of word frequency distribution. Lower score indicates more flat distribution of words and higher diversity.
  \item \textbf{Repetition}. We directly take $H_{rep}$ from Equation \ref{eq:H_rep_1} to evaluate repetition tendency, which reflects \emph{diversity within the sample}. Higher score indicates less repetition and higher diversity.
\end{itemize}



As is shown in Figure \ref{fig:all}\protect\subref{fig:ppl_gpt2} and \ref{fig:all}\protect\subref{fig:ppl_gpt2_xl}, the PPL of generated samples using our algorithm can also \emph{achieve human level perplexity} but \emph{with more strictly filtered vocabulary}, which means our algorithm truncates more low-probability ``tails'' and still achieves equal PPL to human text, which is a desirable feature, since low-probability ``tails'' that contain unreasonable candidates will lower the quality of the generated text. This indicates that our algorithm relieves text degeneration not only by letting in the ``tails'' but also by permutating the ``head'', unlike traditional methods that \emph{solely} rely on the ``tails''.

Note that our algorithm is highly sensitive to filtering metric, which is caused by the fast increase of additional term $m$ from Corollary 1 when loosening the filtering. Such diversity gain will be \emph{destructive} (e.g., for $p>0.9$), because the inverse value in term $m$ will grow too big and ``blow up'' the algorithm. Thus the \emph{intersection points} with human PPL is the reasonable choices for our algorithm.


As is clearly shown in Figure \ref{fig:all}\subref{fig:bleu4_gpt2} and \ref{fig:all}\subref{fig:bleu5_gpt2_xl}, the Self-BLEU scores achieved by our algorithm decrease significantly faster than nucleus sampling, which indicates great diversity gain. Note that it can achieve almost the same score with ``pure sampling'' near $p=0.999$ that represents \emph{highest} diversity in traditional methods. This means that our algorithm achieves significantly higher diversity but with less ``tails''.

\label{sec:5.3}

As is shown in Figure \ref{fig:all}\subref{fig:zipf_gpt2} and \ref{fig:all}\subref{fig:zipf_gpt2_xl}, our algorithm can fit identical Zipf coefficient to human text, while nucleus sampling can't. This indicates that the permutation of our algorithm renders more flat vocabulary distribution (by encouraging sampling on less probable tokens) which is \emph{less concentrated, closer to human text} and is \emph{unable to achieve} by plain stochastic sampling (which always picks high-probability candidates and results in peaked and less diverse usage of vocabulary).
\begin{table*}[hbt!]
\centering
\includegraphics[width=6.3in]{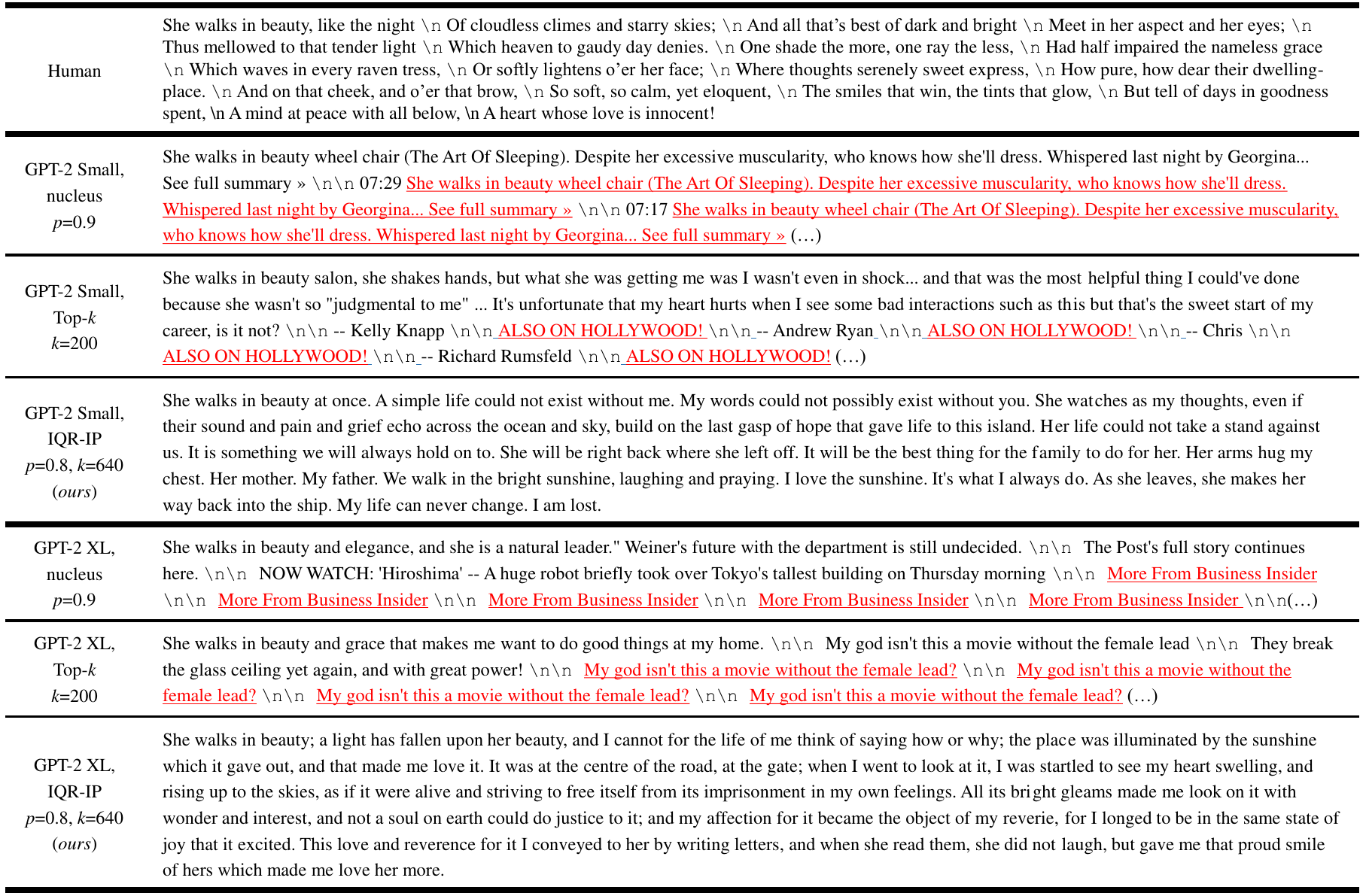}
\caption{Generated examples using different sampling methods that have average perplexity \emph{near human text}. Repetition is marked in red with underline. They show that traditional methods might still generate repetitive sentences, because they only focus on truncating the ``tail'' and ignore the ``head'', while our algorithm generates more diverse, more surprising texts without hurting fluency by permutating the ``head''.}\label{tab:examples_en_1}
\end{table*}

Results for repetition are shown in Figure \ref{fig:all}\subref{fig:entropy_gpt2} and \ref{fig:all}\subref{fig:entropy_gpt2_xl}. Similar to results for Self-BLEU scores, they also show that $H_{rep}$ of our algorithm grows faster and stays higher than nucleus sampling, which represents less repetition and higher diversity.


\subsection{Human Evaluation}

We collect 117 copies of human annotations per each sampling algorithm on \emph{fluency} (focusing on grammar error, linguistic clarity and consistency) and \emph{diversity} (focusing on boredom, wordiness and repetition) on a 1-5 scale (larger better). Results are shown in Table \ref{tab:main_result}. It shows that our algorithm achieves similar fluency score to traditional methods (because they all achieve human-level PPL), which suggests the correct manipulation for the distribution that does not compromise the rationality of the distribution. On the other hand, our algorithm can achieve higher diversity score, which is contributed by the inverse probability permutation that emphasizes on less probable tokens from the ``head''. Note that the corresponding Zipf coefficient of our algorithm indicates \emph{more flat and diverse} distribution of the vocabulary and closer resemblance to human text. And higher $H_{rep}$ of our algorithm indicates \emph{less repetition}. These results clearly suggest diversity gain from our algorithm.

The diversity gain will also reflect on the \emph{style} of the generated text. We present samples from our experiment in Table \ref{tab:examples_en_1}. It can be seen that our algorithm favors creating \emph{diverse and surprising} sentences \emph{without sacrificing fluency}, while traditional method favors creating comparatively \emph{plain and ordinary} sentences. Such difference of language style is also contributed by the inverse probability permutation, which suppresses the sampling for unsurprising high-probability tokens on flat distributions.

\section{Conclusion and Future Work}

In this work we propose the interquartile range inverse probability sampling algorithm. It brings reasonable permutation on the ``head'' of the predicted distribution to enhance diversity without sacrificing fluency. We evaluate our algorithm with pre-trained language models and compare it with traditional stochastic sampling methods. Results show that our algorithm can generate fluent samples with higher diversity and less repetition compared with traditional methods.

Our results reveal a possible direction of \emph{discouraging sampling according to likelihood on flat distributions to increase diversity without hurting fluency}. This might lead to interesting results for other decoding algorithm (such as MIROSTAT, \citealp{basu2021mirostat}) or generation tasks (such as summarization).


\bibliography{poetry2020}
\bibliographystyle{acl_natbib}

\clearpage
\appendix

\section{Proof of Corollary}
\label{sec:appendix_proof}
First, with Pinsker's inequality ~\citep{csiszar_korner_2011}, the total variance between the original filtered distribution $p_{fil}$ and the reference distribution $p_{ref}$ satisfies
\begin{equation}\label{pinsker}
|p_{fil}-p_{ref}|^2\le\frac{1}{2}KL(p_{ref}||p_{fil}).
\end{equation}
Then we may use similar methods by ~\citet{kang-hashimoto-2020-improved} to derive the new bound as follows.
\begin{proof}
\begin{equation}\label{proof1}
\begin{aligned}
|p_{inv}-p_{ref}|^2\le (|p_{inv}-p_{fil}|+|p_{fil}-p_{ref}|)^2
\end{aligned}
\end{equation}
By definition of $p_{inv}$ in Equation \ref{eq:iqr}, we have
\begin{equation}\label{proof2}
|p_{inv}-p_{fil}|^2\le\max_{x\in V^{VeryHigh}}{|p_{fil}-\frac{Z_{p}}{p_{fil}}|}.
\end{equation}
Then expand Equation \ref{proof1}, and use $m$ defined in Equation \ref{eq:proposition1-2} and \ref{proof2} to bound $|p_{inv}-p_{fil}|$, and use Equation \ref{pinsker} to bound $|p_{fil}-p_{ref}|$, the inequality is proved.
\end{proof}
This corollary has the same form as ~\citet{kang-hashimoto-2020-improved}, although with different constant $m$, which corresponds to the truncation ratio $c$ of their proposition. In our work, $m$ is controlled by inverse probability permutation and can be fairly large, while the truncation ratio $c$ satisfies $0\le c\le 1$. In this way, it can be regarded as an extension from proposition by ~\citet{kang-hashimoto-2020-improved} in a different scenario.

Note that since $0<Z_{p}\le 1$, $\max{|p_{fil}-\frac{Z_{p}}{p_{fil}}|}$ can only be achieved on the largest or smallest value of $p_{fil}$ in $V^{VeryHigh}$, i.e., on the first or last candidate of $V^{VeryHigh}$. As a result, $m$ is controlled by $\rho$ in Equation \ref{eq:iqr_div} and filtering parameters in Equation \ref{eq:k1a}. For example, with a loosely filtered $V^{K_1}$, $V^{VeryHigh}$ might contain a last candidate with too small value of probability and render too large value of $m$, hence the total variance will become too high and corrupt the algorithm. However, with carefully chosen parameters, $m$ may provide reasonable variation that enhances diversity and reduces repetition, as is shown in the evaluation results.

\section{Ablation Study}
\label{sec:ablation}
\def\MyWidth{0.065\textwidth}
\begin{table}[hbt!]
\tiny
\centering
\tabcolsep 1mm
\renewcommand{\arraystretch}{1.4}
\begin{center}
\begin{tabular}{
P{0.075\textwidth}
P{\MyWidth}P{\MyWidth}P{\MyWidth}P{\MyWidth}P{\MyWidth}P{\MyWidth}
}
\toprule
Method & PPL & {\tiny Self-BLEU 4} & {\tiny Self-BLEU 5}  & Zipf Coef. & $H_{rep}$ \\
\midrule
\makecell{GPT-2 XL\\IQR-IP\\$p=0.8$\\$k=640$\\$n=100$\\$\rho=1.5$} & 16.77 & 0.47 & 0.29 & 1.17 & 4.45 \\
\cmidrule{1-6}
$\rho=3.0$ & 14.90 & 0.50 & 0.32 & 1.22 & 4.39  \\
$\rho=5.0$ & 12.76 & 0.52 & 0.34 & 1.26 & 4.34  \\
$\rho=10.0$ & 11.57 & 0.53 & 0.36 & 1.39 & 4.30 \\
$\rho=50.0$ & 9.62 & 0.55 & 0.39 & 1.54 & 4.19  \\
\cmidrule{1-6}
$n=10$ & 13.39 & 0.53 & 0.35 & 1.22 & 4.35  \\
$n=50$ & 16.50 & 0.48 & 0.30 & 1.17 & 4.43  \\
$n=200$ & 19.48 & 0.45 & 0.28 & 1.15 & 4.47  \\
$n=1000$ & 20.52 & 0.44 & 0.27 & 1.15 & 4.48  \\

\bottomrule
\end{tabular}
\end{center}
\caption{Ablation study of IQR coefficient and \emph{top1ctrl} filtering for GPT-2 XL.}\label{tab:ablation}
\end{table}

We present ablation study of IQR coefficient and \emph{top1ctrl} filtering in Table \ref{tab:ablation}. Clearly, when $\rho$ in Equation \ref{eq:iqr_div} increases, it shortens the identification range of $V^{VeryHigh}$ hence decreasing the intensity of inverse probability weighting, which leads to more repetition (with higher Self-Bleu score and lower $H_{rep}$), more concentrated distribution of vocabulary (with higher Zipf coefficient), more plain and unsurprising sentences (with lower PPL). As a result, $\rho$ can be used to control the diversity gain that results in style difference. For example, one may need to tune $\rho$ to higher values, if the generated texts seem to lose fluency and have too many obscure sentences (this may be more suitable for artistic generation that requires high diversity and creativity such as poetry or music generation). If $\rho$ is set to infinity, there will be no $V^{VeryHigh}$ and our algorithm will degrade to plain stochastic sampling filtered by Equation \ref{eq:k_and_p} and \ref{eq:k1a} (this may be more suitable for tasks that require high fluency such as summarization or translation).


For the ablation of \emph{top1ctrl} filtering, Table \ref{tab:ablation} clearly shows that loosening $n$ will be harmful, since generated samples will lose quality (with higher PPL). Although this results in less repetition and higher diversity (with lower Self-Bleu score and higher $H_{rep}$), but clearly due to the ``leakage'' of tail described in Section \ref{sec:leakage}, the diversity gain will be destructive which is introduced by candidates with too low probability that interfere with the identification of $V^{VeryHigh}$, which is also reflected by the decrease of Zipf coefficient that represents more flat distribution of vocabulary. On the other hand, small value of $n$ will over-prune the vocabulary, which indirectly decreases the range of $V^{VeryHigh}$ hence decreasing the intensity of inverse probability weighting, resulting in lower diversity and more repetition.

\section{Further Explanations on IQR}
\label{sec:explanation}

A possible concern of IQR is whether it will interfere with peaked distribution that has only a few reasonable candidates (e.g., $1$ or $2$) with high probability in $V^{K_0}$. Note that by definition of IQR, it will only put ``outliers'' in $V^{VeryHigh}$. Clearly, for $V^{K_0}$ with less than $4$ candidates, they will be partitioned among the ``middle part'' of subsets, i.e., symmetrically distributed on $V^{High}$, $V^{Medium}$ and $V^{Low}$. As a result, on highly peaked distribution with only a few ``unquestionably correct'' candidates with high probability in $V^{K_0}$, there will be no $V^{VeryHigh}$ as we have observed, which means that the inverse probability permutation won't work and the algorithm will degrade into plain stochastic sampling. This indicates that IQR can adaptively work on flat distribution and peaked distribution without compromising fluency.

Another issue to clarify is that by the definition of IQR, there should be a $V^{VeryLow}$ that locates symmetrically to $V^{VeryHigh}$ on the identification range. In our experiment we found that this boundary is always below $0$, i.e., $V^{VeryLow}$ is always empty set during IQR calculation. As a result, we omit the narration for $V^{VeryLow}$.

Note that one may even design different and more ``mild'' permutation strategies besides Equation \ref{eq:iqr}, e.g., evenly redistributing $V^{VeryHigh}$, or simply adding some noise on $V^{VeryHigh}$, to achieve a less severe permutation bounded by Equation \ref{proof2}. In that case, our algorithm is actually an extreme case that we completely re-order $V^{VeryHigh}$ with inverse probability which brings significant permutation on the predicted distribution.

\end{document}